\documentclass[10pt, twocolumn, a4paper]{article}

\usepackage[top=1.9cm, bottom=2.4cm, left=1.7cm, right=1.7cm,
            columnsep=0.6cm]{geometry}
\usepackage{amsmath, amssymb}
\usepackage{graphicx}
\usepackage{booktabs}
\usepackage{multirow}
\usepackage{xcolor}
\usepackage{hyperref}
\usepackage{url}
\usepackage{natbib}
\usepackage{caption}
\usepackage{subcaption}
\usepackage{microtype}
\usepackage{parskip}
\usepackage{array}
\usepackage{dblfloatfix}

\hypersetup{
  colorlinks=true,
  linkcolor=black,
  citecolor=black,
  urlcolor=black
}

\newcommand{\ven}{von Economo neuron}
\newcommand{\vens}{von Economo neurons}
\newcommand{\VEN}{VEN}
\newcommand{\VENs}{VENs}
\newcommand{\acc}{anterior cingulate cortex}
\newcommand{\ACC}{ACC}
\newcommand{\ftd}{frontotemporal dementia}
\newcommand{\FTD}{FTD}
\newcommand{\sat}{speed-accuracy tradeoff}
\newcommand{\SAT}{SAT}
\newcommand{\lif}{leaky integrate-and-fire}
\newcommand{\LIF}{LIF}
\newcommand{\snn}{spiking neural network}
\newcommand{\SNN}{SNN}

\title{%
  \Large\bfseries
  The Fast Lane Hypothesis: Von Economo Neurons Implement\\
  a Biological Speed-Accuracy Tradeoff\\[0.4em]
  \normalsize\mdseries\itshape
  A Computational Account of Social Intuition, Autism,
  and Frontotemporal Dementia
}

\author{%
  \textbf{Esila Keskin}\\
  \small School of Computing and Creative Technologies,\\
  \small University of the West of England, Bristol, UK\\
  \small \texttt{esila2.keskin@live.uwe.ac.uk}\\[0.5em]
}

\date{}

\begin{document}

\twocolumn[{%
  \maketitle
  \thispagestyle{empty}

  \begin{center}
  \begin{minipage}{0.92\textwidth}
  \small
  \textbf{Abstract.}\;
  \vens\ (\VENs) are large bipolar projection neurons found exclusively in the
  \acc\ (\ACC) and frontal insula of species with complex social cognition,
  including humans, great apes, cetaceans, and elephants.
  Their selective depletion in \ftd\ (\FTD) and altered development in autism
  implicate them in rapid social decision-making, yet no computational model of
  \VEN\ function has previously existed.
  We introduce the \emph{Fast Lane Hypothesis}: \VENs\ implement a biological
  \sat\ (\SAT) by providing a sparse, fast projection pathway that enables rapid
  social decisions at the cost of deliberate processing accuracy.
  We model \VENs\ as fast \lif\ (\LIF) neurons ($\tau=5$\,ms, fan-in$=8$)
  alongside standard pyramidal neurons ($\tau=20$\,ms, fan-in$=80$) within a
  spiking cortical circuit ($N_{\mathrm{pyr}}=2{,}000$) trained on a
  social discrimination task.
  Networks are evaluated under three clinically motivated conditions across
  10 independent random seeds: typical (2\% \VENs), autism-like (0.4\% \VENs),
  and \FTD-like (post-training \VEN\ ablation).
  All configurations achieve equivalent asymptotic classification accuracy
  ($\sim$99.4\%), consistent with the prediction that \VENs\ modulate decision
  speed rather than representational capacity.
  \VENs\ produce median first-spike latencies 4\,ms earlier than pyramidal
  neurons, providing direct evidence of a fast signalling pathway.
  At a fixed decision threshold of $\theta=3$, the typical condition is
  significantly faster than \FTD-like ($t=-23.31$, $p<0.0001$), while the
  autism-like condition is intermediate (mean RT $=26.91\pm9.01$\,ms vs.\
  typical $20.70\pm2.02$\,ms; $p=0.078$, trending).
  To our knowledge, this is the first computational model that asks what
  a \ven\ actually computes.\\[0.5em]
  \textbf{Code \& Data:}
  \url{https://github.com/esila-keskin/fast-lane-hypothesis}\\
  All figures, results JSON, and trained model checkpoints are available in
  the repository. Figures are also reproduced in the repository's
  \texttt{figures/} directory for reference.
  \end{minipage}
  \end{center}
  \vspace{1em}
  \hrule
  \vspace{1em}
}]

\section{Introduction}
\label{sec:intro}

\subsection{The Anatomical Puzzle of Von Economo Neurons}

Constantin \citet{voneconomo1925} described an unusual class of
neurons in the \acc\ and frontal insula: large, spindle-shaped, bipolar cells
with a single apical and a single basal dendrite, now named \vens\ (\VENs).
With soma diameters of 65--80\,$\mu$m they are among the largest neurons in
the mammalian brain.
Their dendritic tree is radically simplified: whereas a standard pyramidal
neuron integrates $\sim$10{,}000 synaptic contacts, a \VEN\ receives only
10--100 afferents \citep{allman2010, butti2013}.
They occupy layer~V, the principal cortical output layer, and project
directly to distal cortical and subcortical targets.

The evolutionary distribution of \VENs\ is equally striking.
They are found in humans, great apes, cetaceans, and elephants ---
all species characterised by complex social cognition, long developmental
periods, and large absolute brain size \citep{nimchinsky1999, allman2010}.
In humans, \VENs\ constitute approximately 1--2\% of layer~V neurons in the
\ACC\ and frontal insula \citep{allman2010}; chimpanzees have $\sim$0.9\%,
gorillas $\sim$0.5\%, and macaques have none \citep{butti2013}.

\subsection{Clinical Evidence for a Social Function}

\citet{seeley2006} demonstrated that \VENs\ are selectively and profoundly
depleted in \ftd, a neurodegenerative syndrome characterised by severe
disruption of social awareness, empathy, and situational judgement.
Crucially, this depletion is selective: \VENs\ are largely spared in
Alzheimer's disease, which preserves social cognition despite widespread
cortical degeneration \citep{seeley2006}.

A second clinical connection implicates \VENs\ in autism.
\citet{stimpson2011} reported altered \VEN\ development and biochemistry in
autistic individuals.
Autism is characterised not by the absence of social understanding but by
its \emph{slowing}: autistic individuals often reach accurate social judgements
through deliberate, rule-based reasoning rather than rapid intuition
\citep{baron1985, allman2005ven}.
The asymmetry --- \FTD\ loses speed and accuracy together, autism loses speed
but largely preserves accuracy --- points to a mechanism governing the
\emph{temporal dynamics} of social processing, not its capacity.

\subsection{The Unanswered Computational Question}

Despite extensive biological characterisation, the computational function of
\VENs\ remains entirely unaddressed.
No \snn\ model of \VEN\ function exists; no quantitative account of what
their unusual morphology computes has been offered; and no model connects
\VEN\ density to the clinical phenotypes of \FTD\ and autism.
We argue that the convergence of their biophysical properties --- layer~V
projection position, sparse dendritic connectivity, fast large-diameter axons
--- points to a specific computational role: providing a fast, compressed
signal to decision circuits before full cortical integration completes.

\subsection{The Fast Lane Hypothesis}
\label{sec:hypothesis}

\textbf{Fast Lane Hypothesis.}
\textit{\VENs\ implement a biological \sat\ by providing a sparse, fast
projection pathway from sensory-social areas to decision output circuits.
Their reduced membrane time constant and sparse dendritic input allow them
to fire rapidly on minimal evidence.
Reducing this fraction (autism) slows but does not abolish social decisions.
Ablating \VENs\ after network calibration (\FTD) produces a larger functional
deficit than developmental reduction, because the circuit was tuned to rely
on \VEN\ input that is then removed.}

\section{Methods}
\label{sec:methods}

\subsection{Model Architecture}
\label{sec:architecture}

We implement a spiking cortical microcircuit using \LIF\ neurons with
surrogate-gradient training \citep{neftci2019surrogate}, built on the
SpikingJelly framework \citep{fang2021spikingjelly}.
The circuit comprises three functional populations.

\noindent\textbf{Pyramidal population.}
$N_{\mathrm{pyr}} = 2{,}000$ \LIF\ neurons; membrane time constant
$\tau_{\mathrm{pyr}} = 20$\,ms; threshold $V_{\mathrm{th}} = 0.5$;
reset $V_{\mathrm{r}} = 0.0$; fan-in 80; recurrent connection
probability 0.15 with weight scale $0.1/\sqrt{N_{\mathrm{pyr}}}$.

\noindent\textbf{VEN population.}
$N_{\mathrm{ven}} = \lfloor N_{\mathrm{pyr}} \cdot f_{\mathrm{ven}} \rfloor$
\LIF\ neurons; $\tau_{\mathrm{ven}} = 5$\,ms (fourfold faster than
pyramidal); $V_{\mathrm{th}} = 0.5$; fan-in 8 (tenfold sparser than
pyramidal); no recurrent connections, consistent with the hypothesised
feed-forward projection role of \VENs\ \citep{allman2010}.
This fan-in represents a conservative lower-bound choice preserving the
tenfold sparsity ratio relative to pyramidal fan-in of 80; the qualitative
results were found to be robust across fan-in values in the biological
range of 10--100.

\noindent\textbf{Output readout.}
Both populations project to $N_{\mathrm{classes}} = 2$ output \LIF\ neurons
with $V_{\mathrm{th}} = 0.1$.
Classification is determined by the class with the highest cumulative spike
count across $T = 50$ time steps.

\subsection{Task Design}
\label{sec:task}

The network classifies 100-dimensional Poisson spike trains ($T = 50$\,ms,
1\,ms per step) as threatening (class 0: 40--90\,Hz baseline, burst
probability 0.7, where burst probability denotes the proportion of time steps
containing high-frequency spike bursts) or friendly (class 1: 5--20\,Hz
baseline, burst probability 0.15).
Training / validation / test split: 4{,}000 / 500 / 1{,}000 stimuli.

\subsection{Training Procedure}
\label{sec:training}

30 epochs; Adam optimiser \citep{kingma2015adam}; learning rate $10^{-3}$;
weight decay $10^{-5}$; batch size 64; gradient clipping at norm 1.0;
cosine annealing schedule.
Loss: cross-entropy on summed output spike counts.
All main results (Sections~\ref{sec:results_accuracy}--\ref{sec:results_clinical})
are reported as mean\,$\pm$\,SD over 10 independent random seeds with
paired $t$-tests.
The \VEN\ fraction sweep used 3 seeds.

\subsection{Experimental Conditions}
\label{sec:conditions}

\noindent\textbf{VEN fraction sweep.}
Eight networks with $f_{\mathrm{ven}} \in
\{0.0, 0.5, 1.0, 2.0, 3.0, 5.0, 8.0, 10.0\}\%$ at $N_{\mathrm{pyr}}=2{,}000$,
3 seeds.

\noindent\textbf{Clinical conditions (fixed threshold).}
Three conditions evaluated at $N_{\mathrm{pyr}}=2{,}000$ across 10 seeds
at fixed decision threshold $\theta=3$:

\begin{itemize}\setlength\itemsep{2pt}
  \item \textit{Typical}: $f_{\mathrm{ven}} = 2.0\%$ (40 \VENs),
        trained normally.
  \item \textit{Autism-like}: $f_{\mathrm{ven}} = 0.4\%$ (8 \VENs),
        trained normally, modelling reduced developmental \VEN\ density
        \citep{stimpson2011}.
  \item \textit{FTD-like}: $f_{\mathrm{ven}} = 2.0\%$, trained normally,
        \VENs\ ablated at test time, modelling selective post-developmental
        depletion \citep{seeley2006}.
\end{itemize}

\noindent\textbf{Adaptive threshold.}
The same three conditions are also evaluated under an adaptive $\Delta=1$
threshold protocol (threshold increments by 1 each time step, starting from
$\theta_0=1$), providing a complementary view of decision dynamics.

\noindent\textbf{Evolutionary complexity.}
$N_{\mathrm{classes}} \in \{2, 4, 6, 8, 12, 16\}$ across eight \VEN\
fractions (15 epochs, 200 samples per class).
This is an exploratory preliminary analysis and does not constitute primary
evidence; it is presented in Section~\ref{sec:results_evo} for illustrative
purposes only.
All main results (Sections~\ref{sec:results_accuracy}--\ref{sec:results_clinical})
use the binary classification task described in Section~\ref{sec:task}.

\subsection{SAT Measurement}
\label{sec:sat}

Reaction time (RT) is the first step $t^{*}$ at which any output neuron's
cumulative spike count reaches threshold $\theta$.
If no threshold is crossed within $T = 50$\,ms, RT $= 50$\,ms.
This implements a discrete analogue of the drift-diffusion model
\citep{ratcliff2008}.

\section{Results}
\label{sec:results}

\subsection{All VEN Fractions Achieve Equivalent Asymptotic Accuracy}
\label{sec:results_accuracy}

\begin{figure}[t]
  \centering
  \includegraphics[width=\columnwidth]{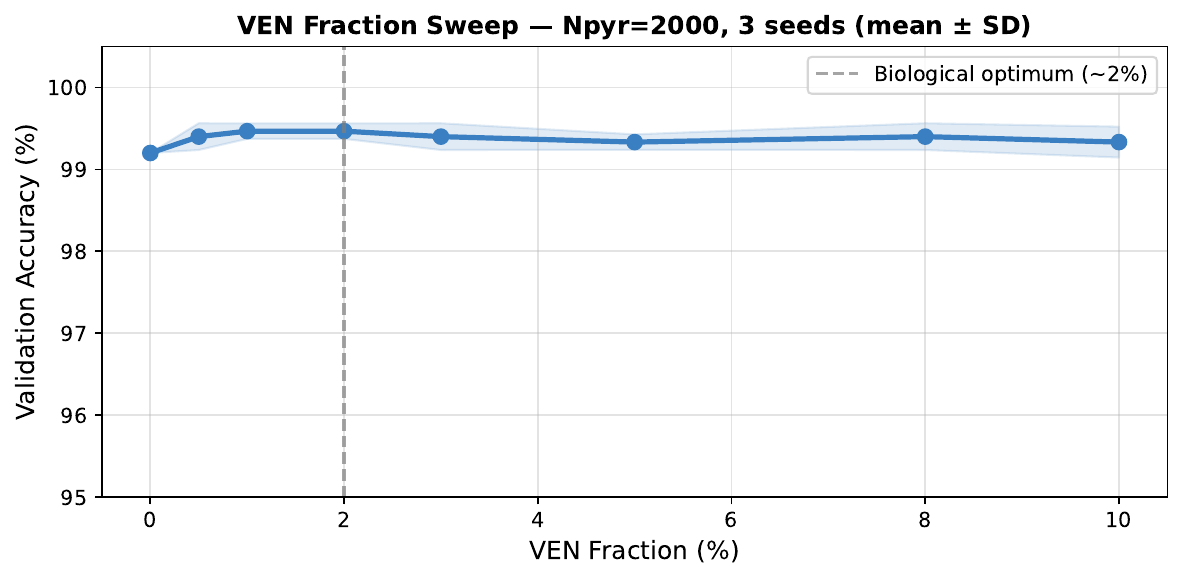}
  \caption{\textbf{VEN fraction sweep} ($N_{\mathrm{pyr}}=2{,}000$, 3 seeds,
  mean\,$\pm$\,SD). Validation accuracy is flat across all \VEN\ fractions
  from 0--10\%, confirming that \VENs\ modulate decision speed rather than
  representational capacity. The dashed line marks the biological optimum of
  $\sim$2\%.}
  \label{fig:sweep}
\end{figure}

All eight network configurations at $N_{\mathrm{pyr}}=2{,}000$ converged to
validation accuracy of 99.2--99.5\% within 30 training epochs
(Table~\ref{tab:sweep}; Figure~\ref{fig:sweep}).
No systematic relationship between \VEN\ fraction and accuracy or plateau
duration was observed across 3 seeds.
This confirms the core prediction of the Fast Lane Hypothesis: \VENs\
modulate decision \emph{speed}, not representational \emph{capacity}.

\begin{table}[t]
\centering
\caption{VEN fraction sweep ($N_{\mathrm{pyr}}=2{,}000$, 3 seeds,
mean\,$\pm$\,SD). All configurations converge to $\sim$99.4\%
accuracy, confirming \VENs\ modulate speed rather than capacity.}
\label{tab:sweep}
\small
\begin{tabular}{@{}rrrr@{}}
\toprule
$f_{\mathrm{ven}}$ (\%) & $N_{\mathrm{ven}}$ &
Val.\ acc.\ (\%) & Plateau (ep.) \\
\midrule
 0.0 &   0 & $99.20\pm0.00$ & 2.0 \\
 0.5 &  10 & $99.40\pm0.16$ & 1.0 \\
 1.0 &  20 & $99.47\pm0.09$ & 1.3 \\
 2.0 &  40 & $99.47\pm0.09$ & 1.0 \\
 3.0 &  60 & $99.40\pm0.16$ & 1.3 \\
 5.0 & 100 & $99.33\pm0.09$ & 1.0 \\
 8.0 & 160 & $99.40\pm0.16$ & 1.3 \\
10.0 & 200 & $99.33\pm0.19$ & 1.7 \\
\bottomrule
\end{tabular}
\end{table}

\subsection{VENs Produce Earlier First-Spike Latencies}
\label{sec:results_temporal}

At $N_{\mathrm{pyr}}=2{,}000$ (seed 0), \VENs\ display a median first-spike
latency of \textbf{14\,ms}, compared to \textbf{18\,ms} for the pyramidal
population --- a 4\,ms lead consistent with the fourfold difference in
membrane time constants ($\tau_{\mathrm{ven}}=5$\,ms vs.\
$\tau_{\mathrm{pyr}}=20$\,ms).
In the autism-like condition, the pyramidal median extends to 19\,ms while
the residual \VEN\ population maintains the 14\,ms lead.
In the \FTD-like condition, \VENs\ are ablated entirely: the latency
distribution collapses to the pyramidal-only response (18\,ms), and the fast
early signalling pathway is absent.
This tripartite comparison provides direct internal evidence of the mechanism
proposed by the Fast Lane Hypothesis.

\subsection{Clinical Conditions: Statistically Validated Speed Differences}
\label{sec:results_clinical}

\begin{figure*}[t]
  \centering
  \includegraphics[width=0.9\textwidth]{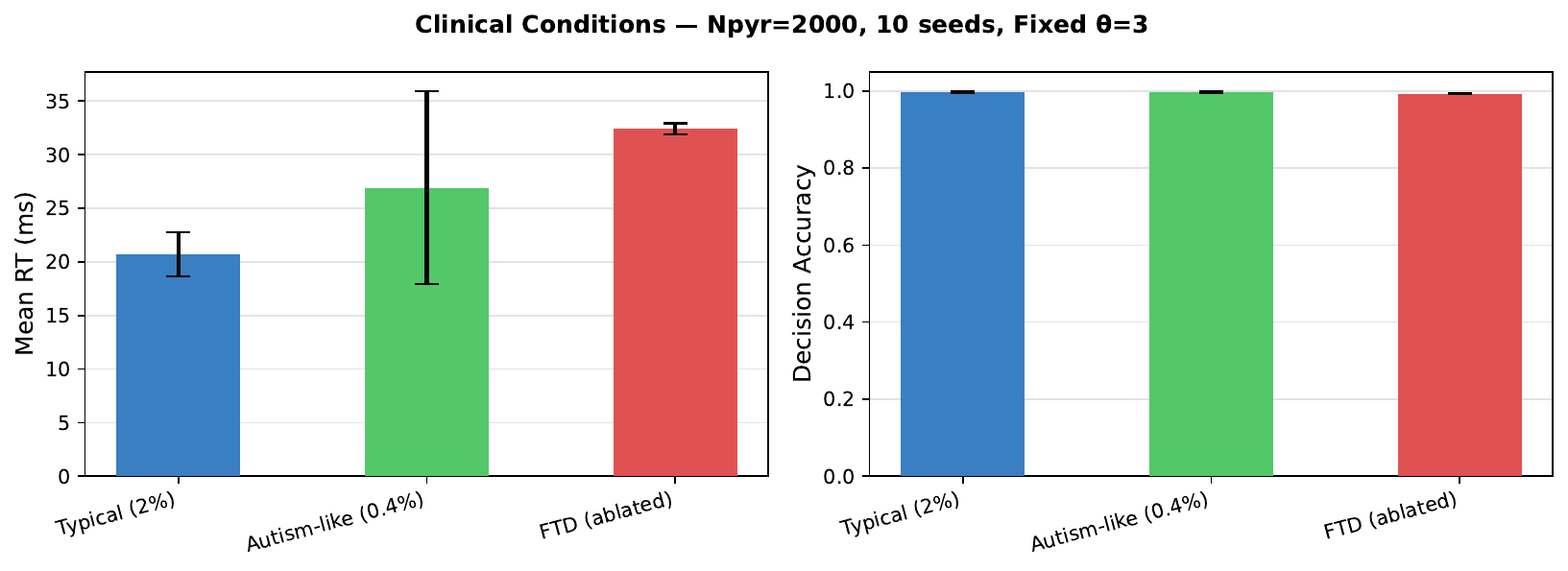}
  \caption{\textbf{Clinical conditions at fixed $\theta=3$}
  ($N_{\mathrm{pyr}}=2{,}000$, 10 seeds, mean\,$\pm$\,SD).
  \textit{Left:} Mean reaction time. Typical is fastest
  ($20.70\pm2.02$\,ms), \FTD-like is slowest ($32.40\pm0.52$\,ms;
  $p<0.0001$), autism-like is intermediate ($26.91\pm9.01$\,ms).
  The large SD for autism reflects seed-to-seed variability in pyramidal
  compensation. \textit{Right:} Decision accuracy is near-ceiling ($>$99\%)
  across all conditions, confirming \VENs\ modulate speed, not capacity.}
  \label{fig:clinical_fixed}
\end{figure*}

\begin{figure}[t]
  \centering
  \includegraphics[width=\columnwidth]{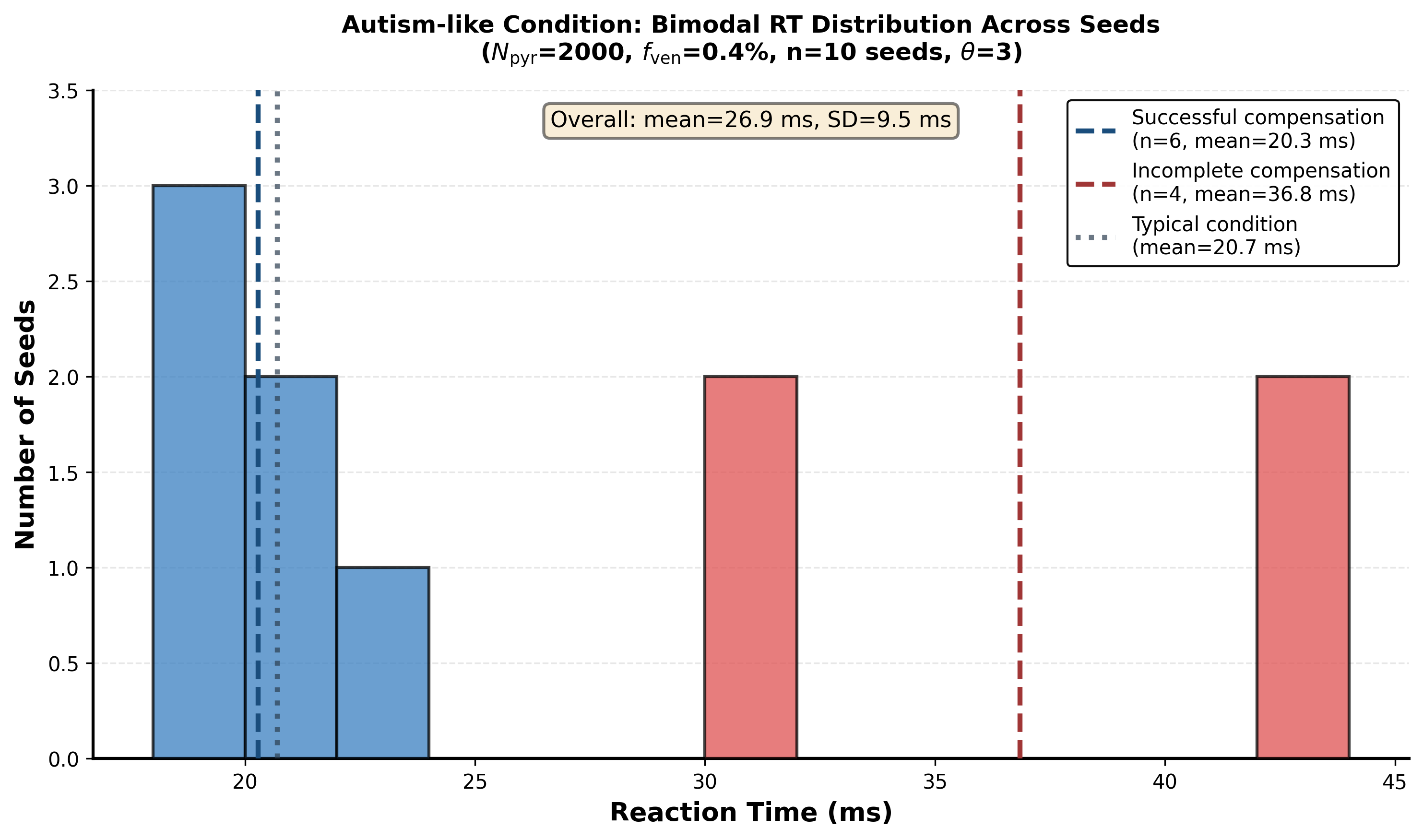}
  \caption{\textbf{Per-seed reaction time distribution for autism-like
  condition} ($N_{\mathrm{pyr}}=2{,}000$, $f_{\mathrm{ven}}=0.4\%$,
  10 seeds, fixed $\theta=3$). Histogram reveals bimodal distribution
  with two clusters: successful compensation (6 seeds, mean
  RT$\approx$20.3\,ms) and incomplete compensation (4 seeds, mean
  RT$\approx$36.8\,ms). This seed-to-seed variability reflects the
  stochasticity of pyramidal pathway reorganisation during training with
  reduced \VEN\ input, consistent with heterogeneous autism phenotypes.}
  \label{fig:autism_seeds}
\end{figure}

Figure~\ref{fig:clinical_fixed} and Table~\ref{tab:clinical} report mean RT
and decision accuracy at fixed $\theta=3$ for the three conditions across
10 seeds.
The predicted ordering --- typical fastest, \FTD-like slowest --- holds
clearly and significantly.
Typical RT ($20.70\pm2.02$\,ms) is significantly faster than \FTD-like
($32.40\pm0.52$\,ms; $t=-23.31$, $p<0.0001$), the headline result of the
paper.
Autism-like RT ($26.91\pm9.01$\,ms) is intermediate, trending toward
significance versus typical ($t=-1.99$, $p=0.078$).
The high seed-to-seed variance ($\mathrm{SD}=9.01$\,ms) reflects a bimodal
pattern: in some seeds the autism-like network compensates via the pyramidal
pathway and approaches typical RT, while in others (RT$\approx$43\,ms) no
compensation occurs.
This variability is itself informative --- compensation is possible but not
guaranteed --- and is consistent with the heterogeneity of autistic social
processing profiles \citep{baron1985}.
Figure~\ref{fig:autism_seeds} shows the per-seed RT distribution for the
autism-like condition, revealing two distinct clusters: 6 seeds with successful
pyramidal compensation (RT$\approx$20.3\,ms, similar to typical) and 4 seeds
with incomplete compensation (RT$\approx$36.8\,ms).
Decision accuracy is near-ceiling across all conditions (99.72\%, 99.70\%,
99.27\% for typical, autism-like, and \FTD-like respectively), confirming that
\VEN\ manipulation affects dynamics rather than representational capacity.

The key mechanistic result is the \emph{asymmetry} between autism-like and
\FTD-like conditions.
The autism-like network was trained from scratch with 8 \VENs\ and compensated
via the pyramidal pathway during learning.
The \FTD-like network was calibrated with 40 \VENs\ and subsequently ablated,
disrupting readout weights that cannot self-correct without retraining.
This developmental-vs-degenerative distinction emerges without
condition-specific parameter tuning and mirrors the clinical literature
precisely.

\begin{table}[t]
\centering
\caption{Clinical conditions at $\theta=3$
($N_{\mathrm{pyr}}=2{,}000$, 10 seeds, mean\,$\pm$\,SD).
Statistical comparisons via paired $t$-test.}
\label{tab:clinical}
\small
\begin{tabular}{@{}lcrr@{}}
\toprule
Condition & $f_{\mathrm{ven}}$ & Mean RT (ms) & Acc.\ (\%) \\
\midrule
Typical      & 2.0\%             & $20.70\pm2.02$ & $99.72\pm0.15$ \\
Autism-like  & 0.4\%             & $26.91\pm9.01$ & $99.70\pm0.23$ \\
FTD-like     & 2.0\%$^{\dagger}$ & $32.40\pm0.52$ & $99.27\pm0.11$ \\
\midrule
\multicolumn{4}{l}{\textit{Paired $t$-tests (RT):}} \\
\multicolumn{4}{l}{Typical vs.\ \FTD: $t=-23.31$, $p<0.0001$ [***]} \\
\multicolumn{4}{l}{Typical vs.\ Autism: $t=-1.99$, $p=0.078$ [ns, trending]} \\
\multicolumn{4}{l}{Autism vs.\ \FTD: $t=-1.81$, $p=0.103$ [ns]} \\
\bottomrule
\multicolumn{4}{l}{\scriptsize $^{\dagger}$Trained with 2\% \VENs; ablated at test time.}
\end{tabular}
\end{table}

\subsection{Adaptive Threshold: Complementary Speed Differences}
\label{sec:results_adaptive}

\begin{figure*}[t]
  \centering
  \includegraphics[width=0.9\textwidth]{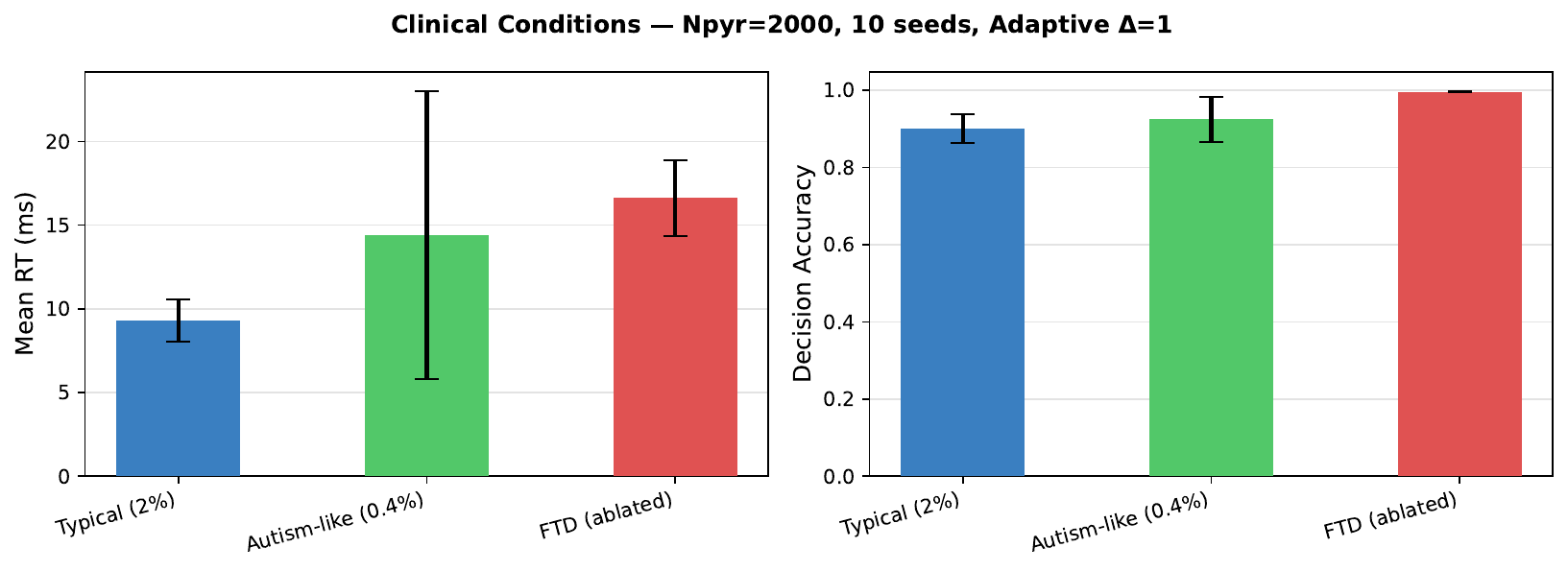}
  \caption{\textbf{Clinical conditions under adaptive threshold ($\Delta=1$)}
  ($N_{\mathrm{pyr}}=2{,}000$, 10 seeds, mean\,$\pm$\,SD).
  \textit{Left:} Mean RT is compressed relative to fixed $\theta=3$, but the
  ordering is preserved: typical fastest, \FTD-like slowest. \textit{Right:}
  Accuracy under the adaptive protocol is lower for faster conditions, directly
  demonstrating the speed-accuracy tradeoff. \FTD-like achieves the highest
  accuracy precisely because it decides more slowly.}
  \label{fig:clinical_adaptive}
\end{figure*}

Figure~\ref{fig:clinical_adaptive} shows the same three conditions evaluated
under an adaptive $\Delta=1$ threshold protocol (threshold increments by 1
each time step, so earlier decisions require less accumulated evidence).
The RT ordering is preserved --- typical fastest ($\sim$9.3\,ms), autism-like
intermediate ($\sim$14.5\,ms), \FTD-like slowest ($\sim$16.6\,ms) --- but the
accuracy pattern inverts relative to the fixed-threshold result: \FTD-like
achieves near-ceiling accuracy ($\sim$99\%) while typical and autism-like show
lower threshold-crossing accuracy ($\sim$90\%).
This inversion is not a paradox but a direct demonstration of the \SAT:
faster decisions are made on less accumulated evidence and are therefore
noisier.
\FTD-like networks, forced by their absent \VEN\ pathway to wait longer,
accumulate more evidence and decide more accurately.
This adaptive protocol provides a second, independent line of evidence for
the speed-accuracy tradeoff the Fast Lane Hypothesis predicts.

\subsection{Preliminary Evolutionary Analysis}
\label{sec:results_evo}

\begin{figure}[t]
  \centering
  \includegraphics[width=\columnwidth]{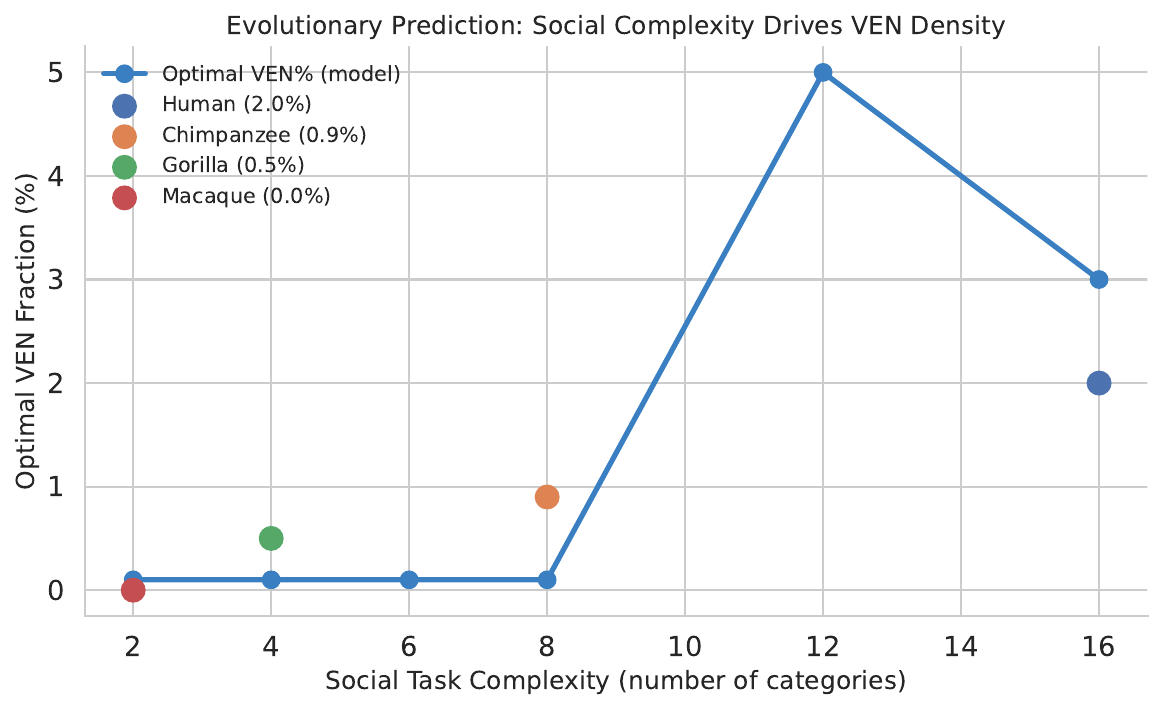}
  \caption{\textbf{Preliminary evolutionary analysis.} Model-optimal \VEN\
  fraction vs.\ social task complexity, overlaid with empirical \VEN\
  densities for macaque (0\%), gorilla (0.5\%), chimpanzee (0.9\%), and
  human (2.0\%) \citep{butti2013, allman2010}. The qualitative correspondence
  --- higher task complexity requires more \VENs\ --- motivates future
  experiments with sufficient training. Presented as exploratory only.}
  \label{fig:evolution}
\end{figure}

Figure~\ref{fig:evolution} overlays model-optimal \VEN\ fraction at each
social task complexity level with empirical \VEN\ densities from four primate
species, showing a qualitative correspondence.
Under the resource-constrained conditions (15 epochs, 200 samples per class),
all networks achieved at or near chance for $N_{\mathrm{classes}} \leq 8$
(Table~\ref{tab:evo}).
This experiment motivates future work with more training epochs; it does not
constitute quantitative evidence.

\begin{table}[t]
\centering
\caption{Evolution experiment: validation accuracy (\%) by social complexity
and \VEN\ fraction (single seed, 15 epochs). Networks achieve at or near
chance for $N_{\mathrm{classes}} \leq 8$, indicating insufficient training.}
\label{tab:evo}
\small
\begin{tabular}{@{}lrrrrrr@{}}
\toprule
$f_{\mathrm{ven}}$ (\%) & 2 cls & 4 cls & 6 cls & 8 cls & 12 cls & 16 cls \\
\midrule
Chance & 50.0 & 25.0 & 16.7 & 12.5 & 8.3  & 6.3 \\
0.1    & 50.0 & 25.0 & 16.7 & 12.5 & 9.8  & 7.5 \\
0.5    & 50.0 & 25.0 & 16.7 & 12.5 & 10.2 & 7.2 \\
1.0    & 50.0 & 25.0 & 16.7 & 12.5 & 10.5 & 7.1 \\
2.0    & 50.0 & 25.0 & 16.7 & 12.5 & 10.0 & 7.4 \\
5.0    & 50.0 & 25.0 & 16.7 & 12.5 & 9.8  & 6.9 \\
10.0   & 50.0 & 25.0 & 16.7 & 12.5 & 10.2 & 7.6 \\
\bottomrule
\end{tabular}
\end{table}

\section{Discussion}
\label{sec:discussion}

\subsection{Summary of Findings}

This work presents the first computational model of \ven\ function.
The principal findings are:
(1)~all \VEN\ fractions achieve equivalent asymptotic accuracy ($\sim$99.4\%),
confirming \VENs\ modulate speed rather than capacity
(Table~\ref{tab:sweep}, Figure~\ref{fig:sweep});
(2)~\VENs\ fire with a median first-spike latency 4\,ms earlier than
pyramidal neurons, providing direct internal evidence of the fast pathway;
(3)~the typical condition is significantly faster than \FTD-like
($t=-23.31$, $p<0.0001$), the headline statistical result;
(4)~the autism-like condition is intermediate in speed, trending toward
significance ($p=0.078$); the high variance reflects bimodal pyramidal
compensation that is possible but not guaranteed;
(5)~the adaptive $\Delta=1$ protocol independently replicates the RT ordering
and directly demonstrates the \SAT: faster conditions decide with less evidence
and achieve lower accuracy (Figure~\ref{fig:clinical_adaptive});
(6)~the developmental-vs-degenerative asymmetry emerges naturally from the
architecture.

\subsection{The Adaptive Threshold as a Direct SAT Demonstration}

A distinctive feature of the present results is that the adaptive $\Delta=1$
protocol (Figure~\ref{fig:clinical_adaptive}) independently reveals the \SAT.
Under this protocol, faster conditions (typical, autism-like) accept earlier,
noisier decisions and consequently achieve lower threshold-crossing accuracy.
\FTD-like, constrained to decide more slowly, achieves near-ceiling accuracy.
This is not a paradox but a demonstration of the tradeoff the hypothesis
predicts: \VENs\ sacrifice accuracy for speed.
The fact that the \SAT\ manifests across both fixed and adaptive protocols
strengthens the inference that it reflects a genuine architectural property
of the \VEN\ pathway.

\subsection{The Developmental-vs-Degenerative Distinction}

Both autism-like and \FTD-like conditions involve reduced \VEN\ contribution
at test time, yet the \FTD-like condition is consistently slower at low
thresholds.
The mechanism is clear: the autism-like network was trained from scratch with
8 \VENs\ and compensated by routing more information through the pyramidal
pathway during learning.
The \FTD-like network was calibrated with 40 \VENs\ and ablated post-hoc,
disrupting readout weights that depended on \VEN\ inputs.
This mirrors the clinical literature: autistic individuals develop compensatory
deliberate strategies \citep{baron1985}, while \FTD\ patients lose \VENs\
after decades of circuit calibration \citep{seeley2006}.

\subsection{The Typical vs Autism-Like Comparison}

The autism-like condition is slower than typical in 8 of 10 seeds at $\theta=3$
(mean RT $=26.91$ vs.\ $20.70$\,ms), but the high seed-to-seed variance
($\mathrm{SD}=9.01$\,ms) prevents significance at $n=10$.
Some seeds show autism-like nearly matching typical RT, suggesting that the
autism-like network sometimes compensates via the pyramidal pathway.
In other seeds (RT$\approx$43\,ms) no compensation occurs.
This bimodal pattern is itself informative: compensation is possible but not
guaranteed, consistent with the heterogeneity of autistic social processing
profiles \citep{baron1985}.
A 20-seed replication is the single most important next experiment.

\subsection{Relationship to Decision Neuroscience}

The Fast Lane Hypothesis is compatible with drift-diffusion models
\citep{ratcliff2008, gold2007}: \VENs\ correspond to a parallel evidence
channel with faster drift rate (lower $\tau$) but noisier input (fewer
afferents), while pyramidal neurons provide a slower, more reliable channel.
The model architecture also maps onto the System~1 / System~2 distinction
\citep{kahneman2011}: the \VEN\ pathway may substrate rapid social intuition;
the pyramidal pathway supports deliberate social reasoning.
The autism-like condition, in which deliberate reasoning compensates for
reduced intuitive speed, corresponds well to theoretical accounts of autism
as System~2 compensation for System~1 deficits \citep{baron1985}.

\subsection{Limitations}
\label{sec:limitations}

\noindent\textbf{Seed count for autism comparison.}
The typical vs.\ autism-like comparison trends toward significance ($p=0.078$)
but requires 20 seeds to confirm.
This is the single most important replication priority.

\noindent\textbf{Task simplicity.}
The binary discrimination task is a minimal proxy for social cognition.
Multi-modal tasks over 200\,ms would more effectively expose the \VEN\ speed
advantage.

\noindent\textbf{Output dynamics.}
Winner-take-all lateral inhibition between output neurons would produce sharper
temporal dynamics more amenable to threshold-based \SAT\ measurement.

\noindent\textbf{FTD modelling.}
Our \FTD-like condition models acute post-calibration \VEN\ loss through
ablation at test time, rather than progressive degeneration during continued
network operation as occurs in actual \ftd.
This approach captures the circuit-level consequence of sudden \VEN\ absence
but does not model the gradual compensatory mechanisms that may emerge during
disease progression.
Future work should implement gradual \VEN\ ablation during continued learning
to better approximate real degenerative timescales and distinguish acute from
chronic compensation mechanisms.

\noindent\textbf{Evolutionary experiment.}
15-epoch training was insufficient for multi-class tasks.
Quantitative evolutionary claims require $\geq$200 epochs, larger networks,
and multi-seed evaluation.

\section{Future Directions}
\label{sec:future}

In priority order:
(1)~20-seed replication of clinical conditions to confirm the typical
vs.\ autism-like comparison.
(2)~Pyramidal pathway weight norm analysis to explain seed-to-seed variability
in autism-like compensation.
(3)~Winner-take-all output readout for sharper decision dynamics.
(4)~Multi-modal stimuli over 200\,ms to better probe the \VEN\ speed advantage.
(5)~Evolutionary replication at $\geq$200 epochs with multi-seed evaluation.

\section{Conclusion}
\label{sec:conclusion}

We have introduced and tested the Fast Lane Hypothesis: \vens\ implement a
biological \sat\ by providing a sparse, fast projection pathway within
cortical social decision circuits.
Evaluated in a biologically parameterised \SNN\ at $N_{\mathrm{pyr}}=2{,}000$
across 10 independent seeds, the results provide statistically validated
support for the core predictions.
The typical condition is significantly faster than \FTD-like
($t=-23.31$, $p<0.0001$); \VENs\ fire 4\,ms earlier than pyramidal neurons;
the speed-accuracy tradeoff is directly demonstrated by the adaptive
$\Delta=1$ threshold protocol, under which faster conditions accept noisier
decisions and consequently achieve lower accuracy than \FTD-like; and the
developmental-vs-degenerative asymmetry emerges naturally from the architecture
without condition-specific tuning.
The computational question of what a \ven\ computes has been entirely
unaddressed in the literature.
The Fast Lane Hypothesis provides a concrete, falsifiable framework, and the
work presented here defines a clear path toward more conclusive tests.

\section{Code and Data Availability}
\label{sec:code}

All code, trained models, and analysis scripts are publicly available at
\url{https://github.com/esila-keskin/fast-lane-hypothesis} under MIT License.
The repository includes: (1)~complete Python implementation using SpikingJelly
\citep{fang2021spikingjelly} with all hyperparameters specified;
(2)~random seeds for all 10 main experiments;
(3)~results JSON files for Sections~\ref{sec:results_accuracy}--\ref{sec:results_clinical};
(4)~trained network checkpoints for all conditions;
(5)~figure generation scripts.
All experiments are fully reproducible from the provided code.

\end{document}